\documentclass[10pt,twocolumn,letterpaper]{article}

\usepackage[pagenumbers]{cvpr} 

\definecolor{cvprblue}{rgb}{0.21,0.49,0.74}
\usepackage[pagebackref,breaklinks,colorlinks,allcolors=cvprblue]{hyperref}
\usepackage{epigraph}
\usepackage{lipsum}
\usepackage{float}
\usepackage{setspace}

\def\paperID{17764} 
\def\confName{CVPR}
\def\confYear{2025}

\title{Is Your World Simulator a Good Story Presenter?\\A Consecutive Events-Based Benchmark for Future Long Video Generation}

\author{
Yiping Wang$^{1}$
\and 
Xuehai He$^{2}$  
\and
Kuan Wang$^{3}$ 
\and 
Luyao Ma$^{4}$ 
\and
Jianwei Yang$^5$
\and
Shuohang Wang$^5$
\and 
Simon Shaolei Du$^1$
\and
Yelong Shen$^5$ 
\and
\\
$^1$University of Washington\quad 
$^2$University of California, Santa Curz\quad
$^3$Georgia Institute of Technology\\
$^4$University of California, San Diego\quad
$^5$Microsoft
\\ 
{\small
\texttt{
\{ypwang61,ssdu\}@cs.washington.edu}
},\quad
\small
\texttt{
\{yelong.shen,shuohang.wang\}@microsoft.com
}
\\
}

\newcommand{\modelname}{StoryEval\xspace}

\begin{document}

\makeatletter
\let\@oldmaketitle\@maketitle
\renewcommand{\@maketitle}{\@oldmaketitle
 \captionsetup{type=figure}
  \vspace{-5.5ex}
  \centering
  \includegraphics[width=0.82\linewidth]{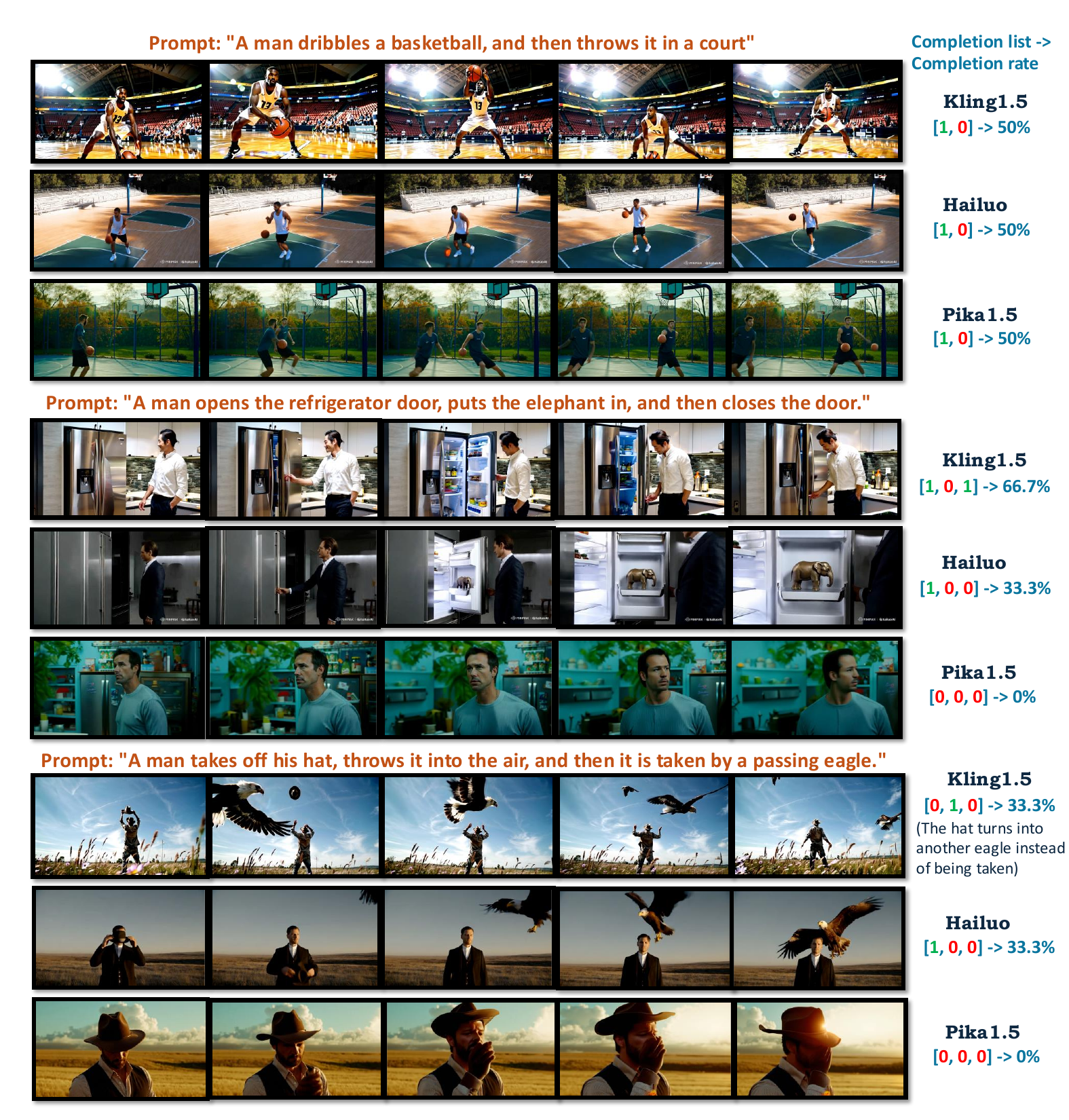}
   \vspace{-3ex}
    \captionof{figure}{
    \textbf{
    Even the top video generative models always fail to completely present some short stories}, such as  ``using 3 steps to place an elephant into the refrigerator''.
    We evaluate some top commercial models (Kling1.5 (10s), Hailuo (6s), and Pika1.5 (6s)) on our \modelname,  whose prompts contain short stories composed of several consecutive events. ``completion list'' denotes if each event is completed (0/1: in-completed/completed) in the generated videos, and ``completion rate'' takes its average.
    For example, in the first prompt, all three models show the man dribbles the basketball, but none of them shows throwing ball, so completion list = [1,0] and completion rate = 50\%.
    }
    \label{fig:show}}

\maketitle

\footnotetext{This work was done during Yiping's internship at Microsoft.}

\begin{abstract}









The current state-of-the-art video generative models can produce commercial-grade videos with highly realistic details. However, they still struggle to coherently present multiple sequential events in the stories specified by the prompts, which is foreseeable an essential capability for future long video generation scenarios. 
For example, top T2V generative models still fail to generate a video of the short simple story "how to put an elephant into a refrigerator." 
While existing detail-oriented benchmarks primarily focus on fine-grained metrics like aesthetic quality and spatial-temporal consistency, they fall short of evaluating models' abilities to handle event-level story presentation.
To address this gap, we introduce \textbf{StoryEval}, a story-oriented benchmark specifically designed to assess text-to-video (T2V) models' story-completion capabilities. StoryEval features 423 prompts spanning 7 classes, each representing short stories composed of 2–4 consecutive events. We employ Vision-Language Models, such as GPT-4V and LLaVA-OV-Chat-72B, to verify the completion of each event in the generated videos, applying a unanimous voting method to enhance reliability. Our methods ensure high alignment with human evaluations, and the evaluation of 11 models reveals its challenge, with none exceeding an average story-completion rate of 50\%. StoryEval provides a new benchmark for advancing T2V models and highlights the challenges and opportunities in developing next-generation solutions for coherent story-driven video generation.
Refer to our  \href{https://ypwang61.github.io/project/StoryEval/index.html}{project page} for more details.
\end{abstract}    
\vspace{-0.5em}

\begin{quote}
\raggedleft
``The universe is made of stories, not of atoms.'' \\
--- Muriel Rukeyser
\end{quote}

\section{Introduction}
The advent of Sora~\cite{Sora2024Brooks} has boosted the rapid evolution in the field of video generation, 
demonstrating the potential of video generative models as world simulators~\cite{xiang2024pandora_world_simulaor,meng2024PhyGenBench_world_simulator,qin2024worldsimbench_world_simulator,doubao2024phy_law_per_world_simulator,wang2024worlddreamer_world_simulaor}. 
Its exciting achievements enable people to mimic the real or imagined world by generating highly realistic and creative videos using simple text prompts. 
For text-to-video (T2V) generation, which is one of the central topics in this area, 
advanced open-source diffusion-based generative models are now capable of generating detailed and continuous short videos ranging from 4 to 10 seconds~\cite{easyanimate2024xu,opensora2024zheng,opensoraplan2024PKU_Yuan,videocrafter2_2024chen,yang2024cogvideox,vchitect2024}. 
Moreover, the leading closed-source models can produce commercial-grade examples with significantly better quality~\cite{kling2024,hailuo2024,gen3runwayml2024,pika2024}. 
Beyond the success of scaling video diffusion models~\cite{peebles2023DiT}, recent research has also revealed the impressive potential of autoregressive video generation. 
This approach uses next-token prediction loss to train models, similar to language models, by tokenizing text, images, and videos into a unified discrete space~\cite{weissenborn2019scaling_autoregressive_vg,wang2024loong_autoregressive_vg,Weng_2024_CVPR_ART_V_autoregressive_vg,moviegen2024meta,jin2024pyramid_flow,wang2024emu3}.

\begin{figure*}[ht]
    \centering
    \includegraphics[width=\linewidth]{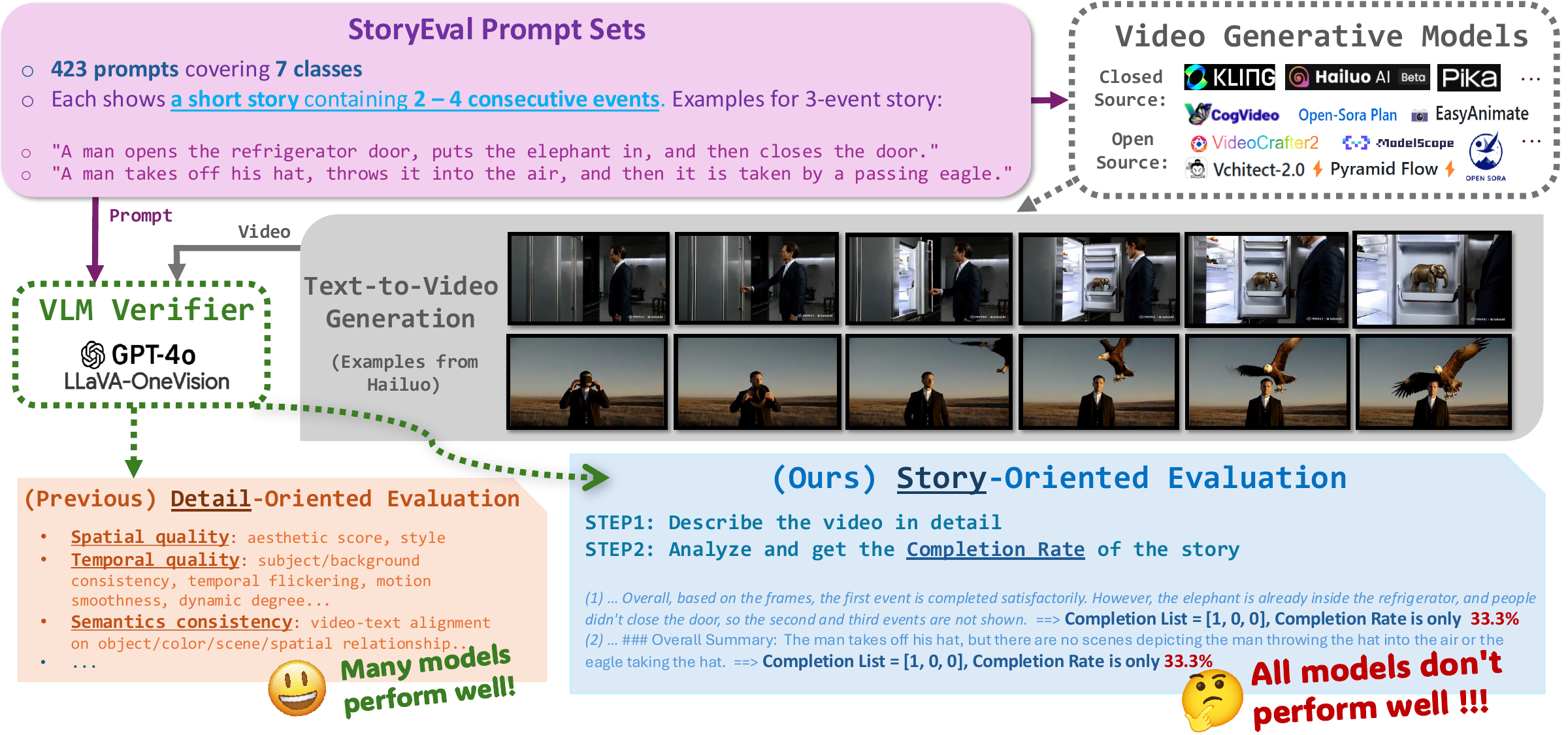}
    \vspace{-1.9em}
    \caption{\textbf{Pipeline of \modelname evaluation.} We carefully design 423 prompts across 5 classes, and each prompt illustrates a short story containing 2-4 sequential events like Figure~\ref{fig:show}. For evaluation, we choose 3 top closed-source commercial models and 8 well-known open-source models, use them for text-to-video generation, and then combine the generated videos and the original prompts as input for VLM verifiers. Different from previous detail-oriented evaluation that focus on fine-grained quality features, we let the VLM to judge how many events are successfully presented in the generated videos, and thus get the completion rate of the story in prompt. Many top models have high performance on previous evaluation, but none of them exceed 50\% completion rate on \modelname.
    }
    \vspace{-0.8em}
    \label{fig:intro_fig}
\end{figure*}

All this progress in T2V generation has stimulated the proposal of various video generation benchmarks~\cite{Ji_2024_T2VBench,huang2024vbench,yuan2024chronomagic_bench,feng2024TC_Bench,liu2024Fetv_Bench,meng2024PhyGenBench_world_simulator,liu2024evalcrafter_bench}. 
However, almost all of these benchmarks focus on evaluating detailed-oriented metrics, which are typically fine-grained (or "low-level") quality features. 
For instance, EvalCrafter~\cite{liu2024evalcrafter_bench} and VBench~\cite{huang2024vbench}, two widely used comprehensive video generation benchmarks, emphasize metrics such as spatial quality (e.g., aesthetic score and styles), temporal quality (e.g., subject/background consistency, temporal flickering, and motion smoothness), and semantic consistency (e.g., video-text alignment in object, color, scene, and spatial relationships), as illustrated under "Detail-Oriented Evaluation" in Figure~\ref{fig:intro_fig}. 
These fine-grained metrics are indeed fundamental indicators of video quality, particularly for short video footage. 
However, due to the rapid advancements in T2V generation, top models are beginning to reach saturation in terms of improvements on these metrics. Moreover, the commercialization of these products demonstrates that humans have already achieved a relatively high level of satisfaction with such details, even as incremental improvements are still expected.

\begin{figure}[t]
    \centering
    \includegraphics[width=0.9\linewidth]{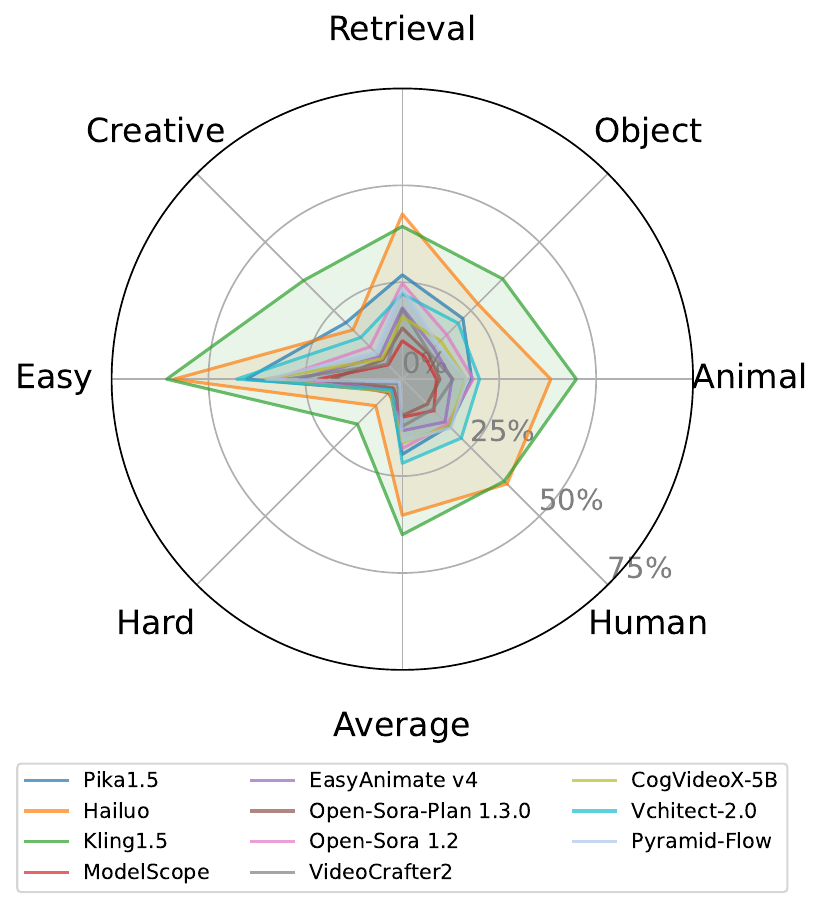}
    \vspace{-1em}
    \caption{\textbf{\modelname evaluation on 11 text-to-video generative models.} 
We visualize their completion rates for the stories across 7 classes, along with the average result for the entire set. 
Even the best model achieves an average completion rate of less than 50\%, meaning it can successfully present fewer than half of the events in a simple short story on average. 
Detailed results are in Table~\ref{tab:all result}.
    }
    \vspace{-1.2em}
    \label{fig:radar}
\end{figure}

On the other hand, long video generation introduces new requirements for evaluation metrics~\cite{wang2023gen_l_video_long_video_generation,henschel2024streamingt2v_long_video_generation,li2024survey_long_video_generation,lu2024freelong_long_video_generation,tan2024video_infinity_long_video_generation}. 
Videos with a longer timespan can include a broader range of temporal variations, often containing several consecutive events rather than a single scenario.
Furthermore, in future industrial applications, users may not only desire aesthetically pleasing individual scenes but also expect to present specific stories in commercials, short videos, or movies on a larger scale. This requires long video generative models to accurately depict all the events within the story prompts. 
For example, if someone wants to create an educational video illustrating \textit{``how to put an elephant into a refrigerator''} and designs a three-step prompt (as shown in the second prompt in Figure~\ref{fig:show}), they would be concerned with whether the generated video accurately presents all three events: ``opens the refrigerator door'', ``puts the elephant in'', and ``closes the door.'' In this context, the \textit{completion rate} of the story, defined as the ratio of events correctly depicted in the generated video, serves as a coarse-grained (or “high-level”) but essential metric for evaluating long video generative models.

Moreover, we surprisingly find that even the current top T2V model  always fail to completely present some simple stories. In Figure~\ref{fig:show}, we see that none of Kling-1.5~\cite{kling2024}, Hailuo~\cite{hailuo2024}, and Pika-1.5~\cite{pika2024} can complete all the events in any short story. 
Consider our previous prompt, the video from Kling1.5 has nothing to do with elephant but just a human opens and closes the refrigerator door, while that from Hailuo shows an elephant sculpture already inside the refrigerator without showing the human action of ``putting in'' and ``closing the door''. Pika generates a nice slow motion shot in kitchen, but the human in video also lacks interaction with refrigerator and elephant. 
Similar phenomena are observed in other prompts, even for those that can be easily retrieved from Internet (as the first prompt).

This gap between future requirements and current model capabilities highlights a meaningful yet under-explored direction for improvement and motivates us to propose a \textit{story-oriented} metric, \textbf{\modelname}, to evaluate a model's capability to present a coherent story. Unlike previous detail-oriented metrics, \modelname focuses solely on the completion rate of consecutive events within a given story, rather than on the quality of fine-grained features.
We summarize our contributions as follows:

\begin{itemize}
    \item We carefully designed and filtered a prompt suite, comprising 423 short stories with 2-4 sequential events. These prompts span 7 categories-—``Human'', ``Animal'', ``Object'', ``Retrieval'', ``Creative'', ``Easy'' and ``Hard''—and are generated using three approaches: retrieving suitable real-world videos and captioning them, human brainstorming, and guided generation with GPT-4o. To ensure a high-quality evaluation process, we further filter some prompts, like those GPT-4o frequently declines to answer due for security reasons, or those have relatively weak alignment with human annotation.
    \item Based on this prompt suite, we design our story-oriented \modelname benchmark. We generate test videos using 3 top commercial closed-source models and 8 well-known open-source models. Then, with VLM verifiers GPT-4o~\cite{gpt4o2024openai} and LLaVA-OV-Chat-72B~\cite{li2024llava_onevision} and the uanimous voting strategy for processing multiple VLM responses, we evaluate the completion rate of the presented story according to carefully designed verifier guidelines. Our results indicate that even the best commercial model achieves less than 50\% completion rate, especially on creative tasks, demonstrating that \modelname is a valuable yet challenging benchmark, and can be an effective complementary metric to the previous detail-oriented metrics.
    \item We further validate \modelname's human alignment on evaluation results, underscoring the reliability of automated scoring. We also conduct some ablation studies about utilizing different VLM verifier and different strategies of applying multiple responses.
\end{itemize}

\section{\modelname}
In this section, we introduce our \modelname in detail. And the related works of the text-to-video generative models and previous benchmarks are shown in Appendix.



\subsection{Prompt Suite}\label{sec: prompt suite}


\subsubsection{How to Create the Story-Oriented Prompt Suite?}\label{sec: construct prompt suite}

\begin{figure}[H]
    \centering
    \includegraphics[width=1.02\linewidth]{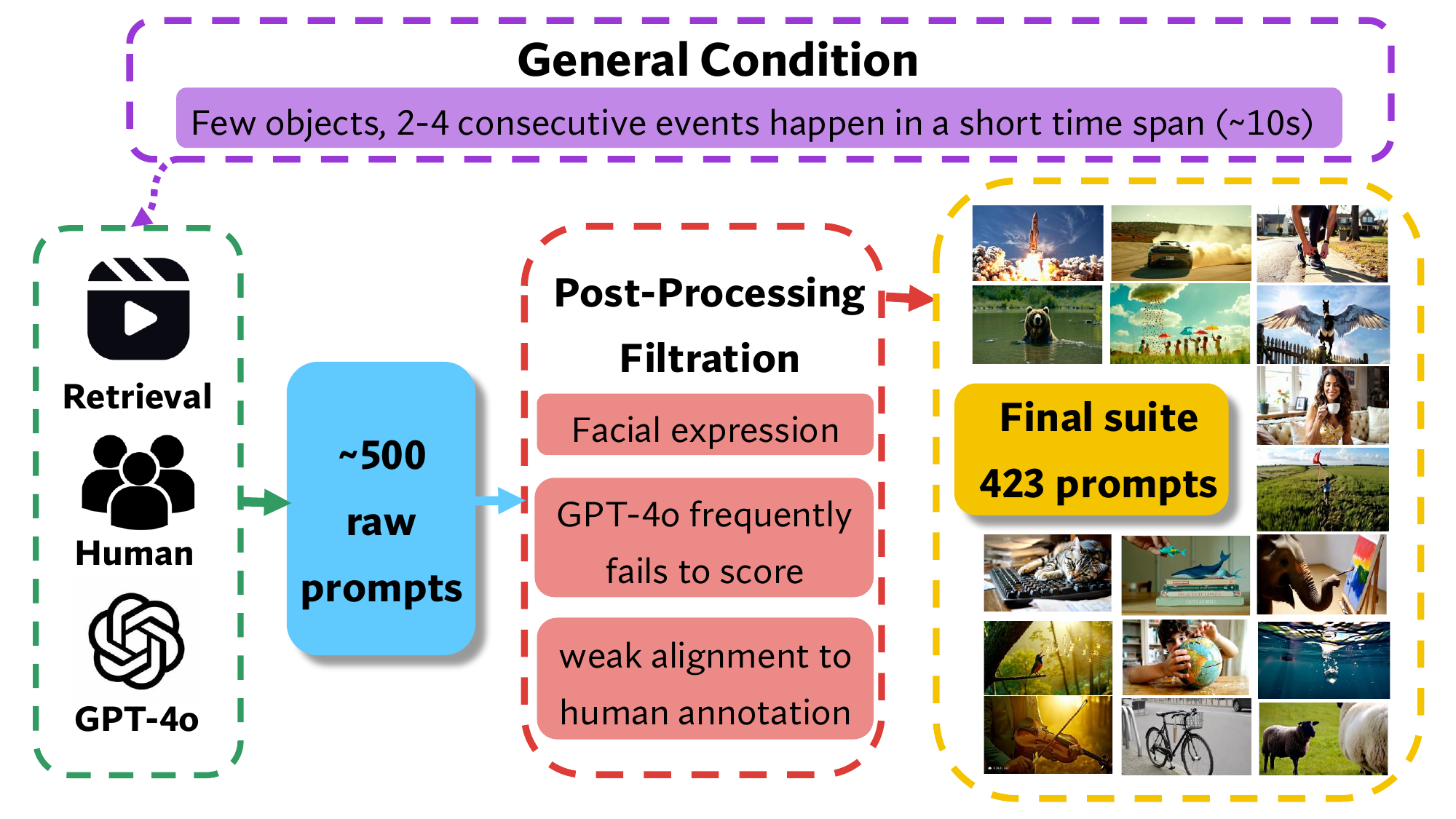}
    \vspace{-2em}
    \caption{\textbf{The process of constructing \modelname prompt suite. } Video examples are selected from three closed-source models: Kling-1.5~\cite{kling2024}, Hailuo~\cite{hailuo2024}, and Pika-1.5~\cite{pika2024}.
    }
    \vspace{-0.5em}
    \label{fig:construct prompt}
\end{figure}

The process of constructing our prompt suite is illustrated in Figure~\ref{fig:construct prompt}.
First, we define some basic conditions that the prompts are expected to satisfy:

\noindent\textbf{General condition.}
To ensure a standard and reasonable evaluation, we aim for \textit{each short story in the prompt suite to involve a small number of physical objects engaging in 2 to 4 consecutive events, and it should be capable of occurring within a relatively short time span so that videos of 10-second-level length can fully present.}
Most of the prompts only consider stories with 1 or 2 subjects, since multi-subject video generation always suffers from issues like missing subjects or assigning actions to incorrect subject~\cite{chen2023videodreamer_multi_subject_t2v_generation,wei2024dreamvideo_multi_subject_t2v_generation,chen2024DisenStudio_multi_subject_t2v_generation,wang2024customvideo_multi_subject_t2v_generation}.
We also exclude movie-like long stories due to the limit 
 of generation length ($\leq$ 10 seconds) for the current T2V models.
Nevertheless, even though these prompts are simple compared to the real-world complex stories, they are still challenging tasks (Figure~\ref{fig:show} and \ref{fig:radar}). 

\begin{figure*}[ht]
    \centering
    \vspace{-1.5em}
    \includegraphics[width=0.35\linewidth]{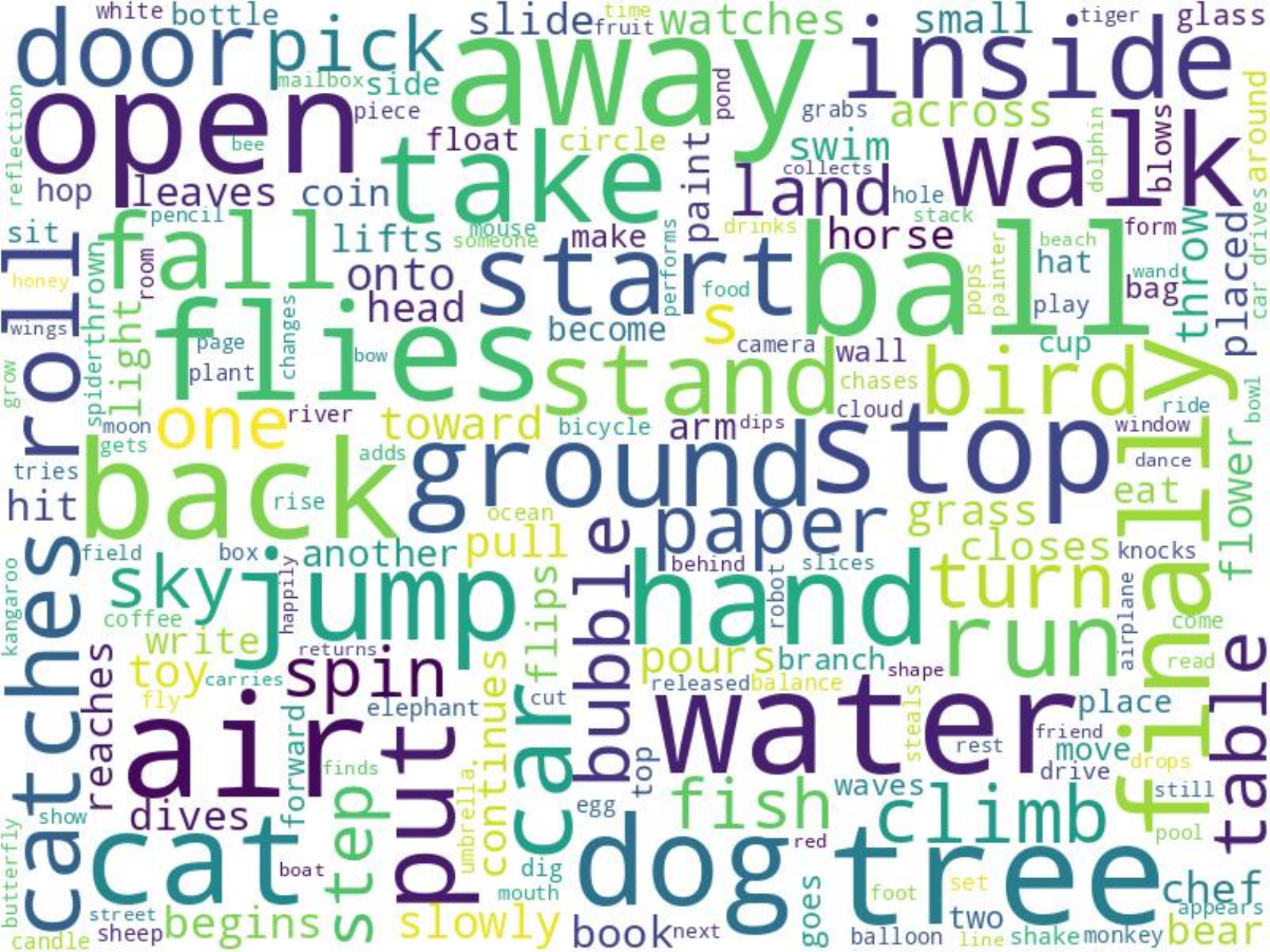}
    \includegraphics[width=0.60\linewidth]{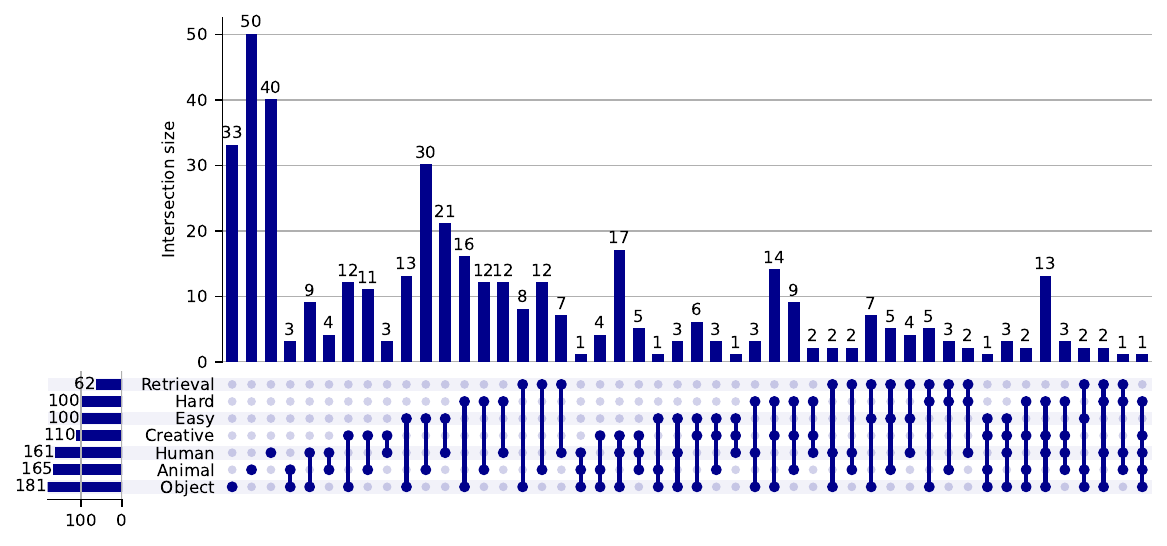}
    \vspace{-0.5em}
    \caption{
    \textbf{Prompt Suite Statistics.} \textbf{(Left)} 
    Word cloud of the StoryEval prompt suite (excluding human-related words like ``person'' to show more diverse terms). \textbf{(Right)} We visualize the proportion of 7 classes in the prompt suite using an UpSet plot. The bottom left of the figure shows the number of prompts in each class, while the right side displays the number of prompts belonging to each class-intersection group. For example, there are 14 examples that exactly belong to all three classes: ``Hard'', ``Creative'', and ``Object''.
    }
    \label{fig: statistics}
\end{figure*}

Then to make a prompt suite that contains short story with diverse topics, we use three different approaches to create the prompts based on the general condition.

\vspace{-0.5em}

\noindent\textbf{Approach 1: Retrieving and captioning the real-world videos.} First, to create prompts that align with real-world distributions, we search the proper videos on Youtube, Pexels~\cite{pexels2024dataset}, and Panda-70M~\cite{chen2024panda70m_dataset}, 
using diverse keywords including different human occupations, different animals, and different non-animate objects like vehicles, etc. 
We selected short video clips that met the general conditions described above and obtain about 70 prompts by manually captioning the videos (like the first ``playing basketaball'' prompt in Figure~\ref{fig:show}). This process naturally implies that \textit{these retrieval-based videos can achieve 100\% completion rate for their prompts in human annotators' verification}.


\noindent\textbf{Approach 2: Human Brainstorming.}  
To create more diverse and imaginative prompts, we also engage in brainstorming and write some examples. In addition to real-world-like prompts, we designe interesting short stories that are logically plausible but highly improbable (e.g., the third prompt in Figure~\ref{fig:show}, where a passing eagle takes a thrown hat). We also include scenarios that are impossible in the real world but can be imagined or presented, like in animated style (e.g., the second prompt in Figure~\ref{fig:show}, where an elephant is put into a refrigerator). 
This approach result in approximately 40 prompts that humans might reasonably expect T2V models to generate, while also providing creative examples tin the third approach.

\noindent\textbf{Approach 3: Guided GPT-4o generation.}
To generate more prompts automatically, we use the previously collected prompts from Approach 1 and 2 as examples, add the general condition as requirements, and combine them to guide GPT-4o for generating more stories.
We also encourage the model to consider diverse keywords of subject or object as Approach 1, and let it cover both the real-world style or creative videos. This approach produce the majority of the prompts, and we carefully select the candidates based on topic/action diversity and matchment on the general condition. As a result, including all three approaches, there are about 500 raw prompts in total.

\subsubsection{Post-Processing Filtration}\label{sec: post filter}
After obtaining the raw prompts, we have some post-processing steps to filter the prompts for better evaluation.
In detail, we will \textbf{filter} the prompts in the following cases:  

    \noindent\textbf{(1) Who uses the facial expression as a event.} 
    For example, if a prompt says someone smiles or cries, we will filter it. The reason is that  GPT-4o will blur the human face in the input, such that it can not judge the facial expressions and will always denote this event as in-completed.
    
    \noindent\textbf{(2) Who makes GPT-4o fails to give the completion rates for more than 2 models\footnote{Here we excludes modelscope~\cite{modelscope2023wang} since it's a very early baseline before Sora, and it fails on more than 50 prompts, which is more than other current advanced models.}.}
    This includes the cases where model fail to generate the videos, like Hailuo sometimes reject to generate the video that shows the faces of children, and the cases where GPT-4o reject to response due to the security concern. 
    Take a prompt that GPT-4o fails on 4 model as an example: ``A chef slices vegetables, and then tosses them into a salad.'', some open-source T2V model will generate videos showing that the knife and hands intertwine with each other, resulting in GPT-4o thinks there is self-harm behavior and refuse to response.
    We filter the cases that let GPT-4o fails on more than two T2V models, while see other failure as the quality issue of generative model itself and counts for zero completion rate for the null response. More details are in Appendix.
    
    \noindent\textbf{(3) Whose corrsponding GPT-4o's evaluation has relatively weak alignment with human feedback.}
    Some of the stories maybe too confusing or complex to GPT-4o, for example, containing events with fast movements or subtle change, such that GPT-4o can not judge whether they are completed.
    We will filter them for ensuring the accurate model ranking. More details are in Sec.~\ref{sec: human annotation}.

After these post-processing filtration, we obtain a final prompt suite containing 423 prompts covering both real-world and imaging examples. 
%











\subsubsection{Prompt Suite Composition}


In \modelname, to obtain richer model results in different downstream tasks, we defined 7 categories based on the participants, sources, styles, and difficulty:

\noindent\textbf{``Human'', ``Animal'' and ``Object''}: For the first three classes, we consider the main participants that lead the story. For instance, in the second prompt in Figure~\ref{fig:show}, both the person and the elephant are the focus of the story, so we classify it into both ``Human'' and ``Animal'' classes. Besides, the ``Object'' here denotes all the non-animate items, like vehicles, buildings, toys, etc.

\noindent\textbf{``Retrieval'' and ``Creative''}:  Another two categories are related to the prompt suite construction process in Sec.~\ref{sec: construct prompt suite}. For the prompts which are the captions of the videos retrieved by real-world (Approach 1), like ``basketball'' example in Figure~\ref{fig:show}, we denote them as ``Retrieval'' class. 
While for the stories that can never happen, like the ``elephant'' example, we assign them as ``Creative'' class.
In the experiment part, we will see that interestingly, most of the models behave relatively better on Retrieval but worse on Creative compared with average result, which may imply that Retrieval examples maybe more similar to the training data of video generative models, while Creative can be out-of-distribution test cases and thus be more challenging.

\noindent\textbf{``Easy'' and ``Hard''}: The final two classes are defined in a post-processing manner. 
We first select the top 8 models which have more than 15\% completion rates averagely in \modelname (As in Table~\ref{tab:all result}).
And then for each prompt, we calculate its average performance over these 8 model, and we define the ``Easy'' class to be top 100 high performance prompts
while ``Hard'' to be the bottom 100 ones.

\noindent\textbf{Statistics.} 
Additionally, we visualize the prompt suite across different categories in Figure~\ref{fig: statistics}. 
There are some interesting patterns from the UpSet plot.
For instance, compared to ``Human'' and ``Animal'', ``Object'' class has a higher relevance to the ``Creative'' and ``Hard'' class, like there are quite a few ``Hard'' examples in ``Object-Creative'' or ``Object-Human-Creative'' intersections. 
Besides, there are 44 ``Creative'' examples also belong to ``Hard'' class, whose ratio (44/110=40\%) is much higher than average (100/423=21\%), meaning that the creative tasks are indeed difficult for most the generative models.








\subsection{Evaluation Process}\label{sec: eval model}

From Figure~\ref{fig:intro_fig}, after obtaining the prompt suite (Sec.~\ref{sec: prompt suite}), we will get some T2V generation examples from the video generative models.
In this section, we will illustrate how we use the VLM verifier to give the completion rates based on these generated videos.


\subsubsection{Evaluation model}

First, we use the current best VLM GPT-4o~\cite{gpt4o2024openai} to obtain the main result, while we also choose one of the best open-source VLMs LLaVA-OV-Chat-72B~\cite{li2024llava_onevision} (LLaVA-OV-Chat-72B) as an additional verifier.
Both these two VLMs use several key frames sampled from the generated examples as input for video understanding. More details are in Appendix.



\subsubsection{Two-Step Quering}\label{sec: two-step querying}
In the evaluation part, we conduct a two-step querying process to get the completion rate. 
In general, this two-step querying process behaves better and more stable than one-step, especially for LLaVA-OV-Chat-72B. 
In the \textbf{first step: describing}, we let the model to describe the given key frames in the video in detail.
Then at the \textbf{second step: scoring}, we still input the key frames of the videos, but additionally append the description obtained from the first step, along with the prompt and the event list, into the second query prompt, in which we let the model judge if each event is completed (mark as 1) or not (mark as 0), and obtain the completion list as shown in Figure~\ref{fig:show} and \ref{fig:intro_fig}.
Here the event list is explicitly defined to prevent the ambiguity about the number of events in the prompt. 
The detailed query prompts can be found in Appendix.

It's worth mentioning that we observe VLM verifiers sometimes provide overly optimistic evaluations to the generated videos, especially for those who are blurry and hard to identify. 
Therefore, we ask VLM to judge the completion of events strictly, resulting in better alignment with human feedback.
More discussions are in Sec.~\ref{sec: majority voting}.
What's more, in the second step, we let the verifer check the \textbf{contextual referent consistency} of the generated videos, meaning that if the prompt implies the subject/object in different events should be the same, but in the video they are different, then verifer should mark the later event as incomplete. For the  ``plaing basketball'' example, if the video shows the man that dribbles the basketball is different from the man that throws the ball, or the basketball that's dribbled is different from the basketball that's thrown, then the later event ``throwing the ball'' will be judged as in-completed.






\subsubsection{Unanimous Voting}\label{sec: majority voting}
The reason for the overestimation phenomenon mentioned in Sec.~\ref{sec: two-step querying} may be that 
VLM is always trained on large-scale real-world videos instead of the synthetic video data, resulting the lack of robustness of VLMs when facing the unclear and noisy generated videos~\cite{chen2024mllm_as_a_judge,chen2024mj_bench,cai2024temporalbench_visual_understanding}.
To solve this issue, we evaluate each prompt-video example multiple times, and then using the voting way for determining the final completion list based on these independent responses.
In detail, assume that for an event in a story, we evaluate it with VLM verifier $N$ times, and then obtain the list of completion score for this event as $\{s_i\}_{i=1}^N$, where each $s_i \in \{0, 1\}$ denotes if the event is considered completed at the $i$-th verification, and the final completion score for this event is denoted as $s^*$. Then the \textbf{$k$-over-$N$ voting} ($k\leq N$) is defined as:
\vspace{-0.5em}
\begin{equation}\label{eq: k-over-N}
\small
    s^* = \mathbb{I}[\sum_{i=1}^N s_i \geq k]
\end{equation}
where $\mathbb{I}(x)$ is the indicator function.
(\ref{eq: k-over-N}) means that $s^*=1$ if and only if there are at least $k$ times the verfier admit that event is completed, otherwise $s^*=0$. 
We use this method in the evaluation process, choose $N=3$ for GPT-4o and $N=2$ for LLaVA-OV-Chat-72B. And we find that the {$N$-over-$N$} voting, which we called \textbf{unanimous voting}, brings the best alignment with human feedback, which is supported by the ablation study in Sec.~\ref{sec: ablation majority}.




\begin{table*}[ht]
\small
  \centering
  \begin{tabular}{@{}l|ccccc|cc|c@{}}
    \toprule
    \textbf{Closed-Source Model} & \textbf{Human} & \textbf{Animal} & \textbf{Object} & \textbf{Retrieval} & \textbf{Creative} & \textbf{Easy} & \textbf{Hard} & \textbf{Average} \\
    \midrule
    Pika-1.5~\cite{pika2024} 
    & 17.1\% & 17.9\% & 22.1\% & 26.9\% & \underline{20.6\%} & 40.3\% & 4.4\% & 19.4\%\\
    Hailuo~\cite{hailuo2024}
    & \textbf{38.2\%} & \underline{38.3\%} & \underline{27.5\%} & \textbf{42.6\%} & 18.0\% & \underline{58.9\%} & \underline{9.7\%} & \underline{35.1\%}\\
    Kling-1.5~\cite{kling2024} 
    & \underline{37.2}\% & \textbf{44.9\%} & \textbf{36.6\%} & \underline{39.4\%} & \textbf{36.0\%} & \textbf{60.8\%} & \textbf{16.4\%} & \textbf{40.1\%} \\
    \midrule
    \textbf{Open-Source Model}\\
    \midrule
    ModelScope~\cite{modelscope2023wang} 
    & 11.5\% & 8.8\% & 8.2\% & 9.9\% & 5.2\% & 21.7\% & 5.0\% & 9.8\%\\
    EasyAnimate v4~\cite{easyanimate2024xu} 
    & 15.6\% & 12.8\% & 11.4\% & 18.4\% & 8.2\% & 26.7\% & 3.6\% & 13.3\% \\
    Open-Sora-Plan 
    1.3.0~\cite{opensoraplan2024PKU_Yuan} 
    & 9.1\% & 9.7\% & 9.4\% & 13.2\% & 7.1\% & 18.2\% & 3.2\% & 9.4\% \\
    Open-Sora 1.2~\cite{opensora2024zheng} 
    & 16.4\% & \underline{18.3\%} & \underline{16.2\%} & \textbf{24.7\%} & \underline{11.8\%} & 32.7\% & \underline{4.3\%} & \underline{17.9\%}\\
    VideoCrafter2~\cite{videocrafter2_2024chen} 
    & 11.0\% & 13.1\% & 9.8\% & 18.0\% & 5.5\% & 29.9\%  & 1.7\% & 12.2\% \\
    CogVideoX-5B~\cite{yang2024cogvideox} 
    & 17.1\% & 16.4\% & 14.0\% & 16.0\% & 7.4\% & \underline{35.4\%} & \textbf{4.6\%} & 16.4\% \\
    Vchitect-2.0~\cite{vchitect2024} 
    & \textbf{21.5\%} & \textbf{19.9\%} & \textbf{20.4\%} & \underline{22.0\%} & \textbf{15.2\%} & \textbf{42.8\%} & 3.9\% & \textbf{21.7\%} \\
    Pyramid-Flow~\cite{jin2024pyramid_flow} 
    & \underline{17.8\%} & 16.5\% & 12.8\% & 23.4\% & 9.7\%  & 35.1\%  & 1.0\% & 16.0\% \\
    \bottomrule
  \end{tabular}
  \vspace{-0.5em}
  \caption{\textbf{{\modelname} evaluation results on 11 video generative models with GPT-4o verifier.}
  Higher the better. None of the models achieves a completion rate higher than 50\%, i.e., successfully present half of the events in the short simple stories on average.
  Notably, ``Retrieval'' class includes the captions that are manually annotated for the videos retrieved from real-world, which means that in human aspect, the upper bound of this metric can be 100\% if the retrieved videos are used as a baseline. More details are in Appendix.
  }
  \vspace{-1em}
  \label{tab:all result}
\end{table*}

\subsection{Human Annotation}\label{sec: human annotation}

At last, we perform human annotation experiment on some generation videos to validate that \modelname's GPT-4o evaluation can align with human's evaluation, both on each class and the entire prompt set. And also we use these annotation for the post-processing filtration as illustrated in Sec.~\ref{sec: post filter}. 

\subsubsection{Data Preparation}

Due to the high cost of evaluating all the videos from all 11 video generative models, for each prompt, we randomly sample 2 different models from 11 candidates, and use their generated videos. These result in about 900 examples for annotation, and each generative model is assigned to about 80 prompts (all before post-processing filtration stage).
To minimize the annotators' bias against the model, we anonymize the corresponding model of the sampled videos, randomly shuffle them to obtain a list of ``video-prompt'' pairs for human annotations.
We also mention that due to the above process, different candidate model will have different $\sim$80 prompts, meaning that the model performance in human annotation experiments may be different to that in the \modelname benchmark which consider the whole same prompt suite. 


\subsubsection{Labeling Process}

In general, we just let human play the role of VLM verifier in the ``video-prompt'' data pair as the second step in Sec.~\ref{sec: two-step querying}.
But here we just use one-step querying process, and directly ask the human annotators to judge if every event in the story is completed by selecting the 0-1 score
We give human annotators the same precautions as VLM,
but do not emphasize strict scoring term, since we find that in general human are already conservative about judging whether the events are completed. We provide about 20 pre-questions for annotators to mark first, and then communicate with them about the details before the large-scale annotations to ensure the high quality of annotation process.


\vspace{-0.3em}
\section{Experiment}
In this section, we evaluate the top video generative models on our \modelname benchmark.
\textbf{Models.} We consider three top closed-source models: Kling-1.5~\cite{kling2024}, Hailuo~\cite{hailuo2024}, and Pika-1.5~\cite{pika2024}, and 8 open-source models: ModelScope~\cite{modelscope2023wang}, EasyAnimate~\cite{easyanimate2024xu}, Open-Sora-Plan~\cite{opensoraplan2024PKU_Yuan}, Open-Sora~\cite{opensora2024zheng}, VideoCrafter2~\cite{videocrafter2_2024chen}, CogVideoX-5B~\cite{yang2024cogvideox}, Vchitect-2.0~\cite{vchitect2024}, and Pyramid-Flow~\cite{jin2024pyramid_flow} (auto-regressive-based). More details about the model are in Appendix.
\textbf{Metrics} In the human alignment test, we use Kendall's $\tau$ coefficient and Spearman's $\rho$ coefficient to measure the rank correlation between two lists, which is higher the better, and $\rho=\tau=1$ means the order of two lists is totally the same.




\subsection{Results}\label{sec: exp result}

\begin{figure*}[h]
    \centering
    \includegraphics[width=0.9\linewidth]{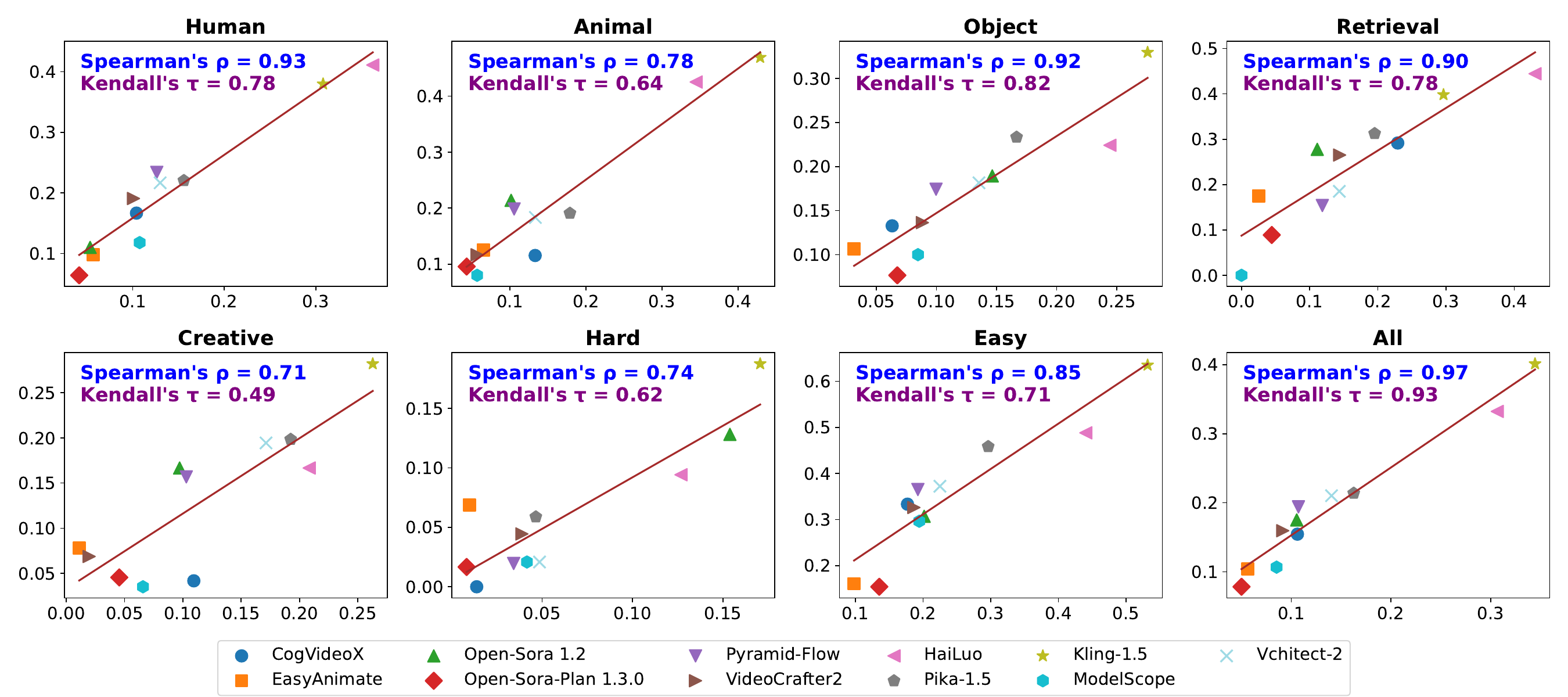}
    \vspace{-0.5em}
    \caption{\textbf{Validate \modelname's human alignment.} In \modelname, the automatic evaluation result obtained by GPT-4o can align the human annotations quite well, especially when considering all the data. 
    Here x-axis denotes average completion rate annotated by human, and y-axis presents that obtained from GPT-4o. We consider the 7 classes and all the data, and calculate the Spearman's and Kendall's coefficient for measuring the rank correlation between the completion rates from human and GPT-4o.
    }
    \label{fig:human-align}
    \vspace{-1.0em}
\end{figure*}

The detailed main results of \modelname benchmark is shown in Table~\ref{tab:all result}. We can obtain the following main observations:

\noindent\textbf{(1) All the current models underperform on \modelname.} 
None of the auto-regressive-based or diffusion-based models we evaluate can achieve 50\% completion rate averagely, which means that complete half of the events in a simple short story. 
Even for the ``Retrieval'' class, which has the retrieved videos as the top baseline with 100\% completion rate, none of them can achieve this. 
If we further consider the most difficult subset ``Hard'', then no models can achieve higher than 20\% completion rate.
This supports our claim that the current models still need a large improvement on story presentation capability.

\noindent\textbf{(2) Closed-source models have large advantage, especially in ``Creative'' tasks, but they are not unbeatable. }
In general, we can see that Kling-1.5 and Hailuo behave much better than other models on all the metrics, while some open-source model can also achieve quite well average performance. For example, Vchitect-2.0 (21.7\%) obtains even better result than Pika-1.5 (19.4\%). 
Additionally, closed-source models behave especially well on ``Creative'' class compare to the open-source ones, which imply that the current closed-source models are more helpful in accomplishing the presentation of those imaginative stories.

\begin{table}
\small
  \centering
  \begin{tabular}{@{}l|cc@{}}
    \toprule
    \textbf{Model (Result (Rank))}  & \textbf{Temporal Consistency} & \textbf{StoryEval} \\
    \midrule
    Kling-1.5 & $\geq$ 97.93\% (1)  & 40.1\% (1)\\
    Pika-1.5 & $\geq$ 97.83\% (2) & 19.4\% (3)\\
    \midrule
    ModelScope & 90.88\% (8)  & 9.8\% (7)\\
    Open-Sora 1.2 &  96.85\% (5)  & 18.2\% (4)\\ 
    Open-Sora-Plan 1.3.0 & 97.28\% ({\color{green}4})  & 9.4\% ({\color{red}8})\\
    VideoCrafter2 & 95.67\% (6)  & 12.2\% (6) \\
    Vchitect-2.0 & 95.03\% ({\color{red}7})  & 21.7\% ({\color{green}2})\\
    Pyramid-Flow & 97.56\% (3)  & 16.0\% (5)\\
    \bottomrule
  \end{tabular}
  \vspace{-0.5em}
  \caption{
  \textbf{\modelname can be an effective complementary metric to the traditional detail-oriented metrics.}
  Temporal consistency result is obtained from VBench~\cite{huang2024vbench}.
  ``$\geq$ X'' means that the VBench records the results X of the previous version of that models (Pika-1.0 (2024-06) and Kling (2024-07 high-performance)), where we simply assume the updated model should be better based on human feedback. 
  Some models can have quite different performance on detail-oriented and story-oriented metrics. 
  }
  \label{tab:vbench compare}
  \vspace{-1em}
\end{table}

\subsubsection{Comparison \modelname with VBench}
Furthermore, we compare the story-oriented metric \modelname together with the traditional detail-oriented one in VBench~\cite{huang2024vbench}.
The result is shown in Table~\ref{tab:vbench compare}. Here the temporal consistency metric includes the evaluation of subject consistency, background consistency, temporal flickering and motion smoothness as in original paper~\cite{huang2024vbench}. 
We can see that some models have quite different performance on these two metrics. For example, Vchitect-2.0 performs good at \modelname but has quite bad temporal consistency, while Open-Sora-Plan 1.3.0 has good VBench results but has the worst story presentation. 
This shows that \textit{maybe models have some tradeoff between temporal consistency and \modelname},  since completing as few events as possible may help the model improve temporal consistency.
More details about VBench results are in Appendix.

\subsubsection{Human Alignment}
As illustrated in Sec.~\ref{sec: human annotation}, we perform human annotation for a subset for the generated videos to validate that our verifier can reflect the human estimation of models' story presenting capability. 
The results are shown in Figure~\ref{fig:human-align}.
We can see that \textit{the completion rates obtained from GPT-4o and human are highly correlated}, which support the reliability of using GPT-4o as VLM verifier.


\subsubsection{Using LLaVA-OV-Chat-72B}
Apart from GPT-4o, we also conduct the experiments using open-source advanced VLM LLaVA-OneVision-Chat-72B as illustrated in Sec.~\ref{sec: eval model}. The results are shown in Table~\ref{tab: llava}. We can observe that the result of using LLaVA-OV-Chat-72B has very close ranking correlations with that of using GPT-4o, showing the robustness \modelname in terms of VLM verifiers.
What's more, we also note that compared to GPT-4o, LLaVA-OV-Chat-72B will further over-estimate the results, resulting higher average completion rates.

\begin{table}
\small
  \centering
  \begin{tabular}{@{}l|cc|c@{}}
    \toprule
    \textbf{Model }  & \textbf{Retrieval} & \textbf{Creative} & \textbf{Average} \\
    \midrule
    Pika-1.5 & 35.2 & \underline{22.9} & 25.0 \\
    Hailuo & \textbf{51.7} & 19.5 & \underline{41.0}\\
    Kling-1.5 & \underline{41.7} & \textbf{30.8} & \textbf{41.7} \\
    \midrule
    ModelScope & 13.7 & 6.2 & 15.2 \\
    EasyAnimate v4 & 21.2 & 10.5 & 18.5 \\
    Open-Sora-Plan 1.3.0 & 17.1 & 6.9 & 12.7 \\
    Open-Sora 1.2 &  \underline{32.2} & \underline{15.4} & \underline{23.6} \\
    VideoCrafter2 & 15.3 & 8.0 & 14.7 \\
    CogVideoX-5B & 27.2 & 8.1 & 19.9 \\
    Vchitect-2.0 & \textbf{33.6} & \textbf{20.5} & \textbf{31.6} \\
    Pyramid-Flow & 26.4 & 10.5 & 20.3 \\
    \midrule
    Spearman's $\rho$ & 0.891 & 0.980 & 0.982\\
    Kendall's $\tau$ & 0.782 & 0.917 & 0.927\\
    \bottomrule
  \end{tabular}
  \vspace{-0.5em}
  \caption{
  \textbf{{\modelname} evaluation results on 11 video generative models with LLaVA-OV-Chat-72B~\cite{li2024llava_onevision} verifier.}
  The $\rho$ and $\tau$ are evaluated between the completion rates of GPT-4o and that of LLaVA-OV-Chat-72B model.
  More details are in Appendix.
  }
  \label{tab: llava}
\end{table}

\subsubsection{Ablation Study of Unanimous Voting}\label{sec: ablation majority}
Finally, we analyze different strategies for applying multiple VLM responses illustrated in Sec.~\ref{sec: majority voting} in Table~\ref{table: vote}. 
We observe that \textit{applying Unanimous voting for GPT-4o can brings the best human alignment result}.
\begin{table}[ht]
    \small
    \begin{tabular}{@{}l|ccc@{}}
    \toprule
        Strategy & Kendall $\uparrow$ & Spearman $\uparrow$ & Mean Diff. $\downarrow$\\
        \midrule
        Vanilla Average & 0.75 & 0.86 & 0.14 \\
        1-over-3 Voting & 0.71 & 0.85 & 0.22 \\
        2-over-3 Voting & 0.71 & 0.86 & 0.14 \\
        \textbf{Unanimous Voting} & \textbf{0.93} & \textbf{0.97} & \textbf{0.05} \\
        \bottomrule
    \end{tabular}
    \vspace{-0.5em}
        \caption{
        \textbf{
        Ablation study about why we use unanimous voting.
        } For each event in every generated video, we have 3 independent completion score from GPT-4o, and then we try vanilla take their average or $k$-over-3 voting as illustrated in Sec.~\ref{sec: majority voting} to process them into a final completion score. Metrics are compared with human annotated completion rates. ``Mean Diff.'' denotes the mean of absolute difference between human score and GPT-4o score over all data and models.
        }
        \vspace{-1em}
        \label{table: vote}
\end{table}


\section{Conclusion}

Our evaluation with StoryEval demonstrates a substantial gap between current video generative models and the demands of coherent story presentation across sequential events. Although top models perform adequately on fine-grained quality measures, they struggle to complete even half of the prompted events, especially within creative scenarios. 
This reveals a critical challenge for advancing T2V models toward real-world applicability in long-format and story-driven video content. 
StoryEval proves to be a reliable and robust benchmark for assessing these limitations, with automated VLM-based scoring well-aligned to human annotations. 
As the field progresses, addressing the specific storytelling and event continuity challenges highlighted by StoryEval will be essential to developing next-generation models.
{
    \small
    \bibliographystyle{ieeenat_fullname}
    \bibliography{main}
}

\clearpage
\maketitlesupplementary

\section{Related Works}
\subsection{Text-to-Video Generative Models}

Recent advancements in vision generation have been largely dominated by
diffusion models ~\cite{peebles2023DiT,easyanimate2024xu,opensora2024zheng,opensoraplan2024PKU_Yuan,videocrafter2_2024chen,yang2024cogvideox,vchitect2024}. The open-source release of Stable Diffusion ~\cite{rombach2022high} has catalyzed extensive research and development in this field, while closed-source models have achieved commercial-grade quality~\cite{kling2024,hailuo2024,gen3runwayml2024,pika2024}. 
In parallel, another promising research direction involves training autoregressive models for video generation, where models are trained to sequentially predict tokens for video generation ~\cite{weissenborn2019scaling_autoregressive_vg,wang2024loong_autoregressive_vg,Weng_2024_CVPR_ART_V_autoregressive_vg,moviegen2024meta,jin2024pyramid_flow,wang2024emu3, marcus2022very, ding2021cogview, yan2021videogpt, kondratyuk2023videopoet}. Despite these advancements, challenges remain in the long-term event continuity and logical narrative progression. 
Some contemporaneous work~\cite{wu2024mind} suggests certain method to alleviate this significant problem, but there is still no standard way to quantify models' capacity in story presentation, and our creative-style prompts can still be challenging for them.
Our benchmark provides a testbed for these lines of research by emphasizing the generation of coherent, long videos grounded in consecutive events. 



\subsection{Video Generation Metrics and Benchmarks}
The rapid advancements in text-to-video (T2V) generation have driven the development of various benchmarks to evaluate the performance of generative models across diverse dimensions~\cite{Ji_2024_T2VBench,huang2024vbench,yuan2024chronomagic_bench,feng2024TC_Bench,liu2024Fetv_Bench,meng2024PhyGenBench_world_simulator,liu2024evalcrafter_bench}. Traditional metrics like FVD \cite{unterthiner2019fvd} and IS \cite{salimans2016improved} have been widely used to assess frame quality and text-frame alignment but fall short in capturing critical aspects such as subject consistency, temporal coherence, and physical commonsense correctness, particularly for novel or dynamic scenes \cite{brooks2022generating}.
Recently, benchmarks like VBench~\cite{huang2024vbench} and EvalCrafter~\cite{liu2024evalcrafter_bench} have expanded the evaluation scope, incorporating advanced metrics for dynamic attributes and human ratings. Specialized benchmarks such as T2V-CompBench~\cite{sun2024t2v} and DEVIL~\cite{liao2024evaluation}, PhyGenBench~\cite{meng2024towards} address compositional and dynamic characteristics and physical realism. However, previous methods largely focus on evaluating detailed dynamics within short videos, failing to address the broader challenge of generating coherent, story-driven long videos. To bridge this gap, our method introduces a holistic evaluation framework that emphasizes narrative completion and the accurate depiction of sequential events, providing a robust benchmark for assessing long video generative models.




\section{Supplementary of \modelname Benchmark}

\subsection{More Details about Prompt Suite}\label{sec: supp prompt suite}
\textbf{Retrieval-Based Prompts.}
Here we claim that the retrieval-based prompts (``Retrieval'' class) we select are designed to match the general condition as illustrated in main paper.
The lengths of videos retrieved from real-world are always hard to control, but the major part of the retrieval-based videos we select cover 5 to 20 second length, and for the videos with longer time span (about 1 minute), their events are also simple and direct enough to be completed in 10 seconds. Some of the retrieved videos are slow shots that can be accelerated. 

\begin{figure*}[t]
    \centering
    \includegraphics[width=0.7\linewidth]{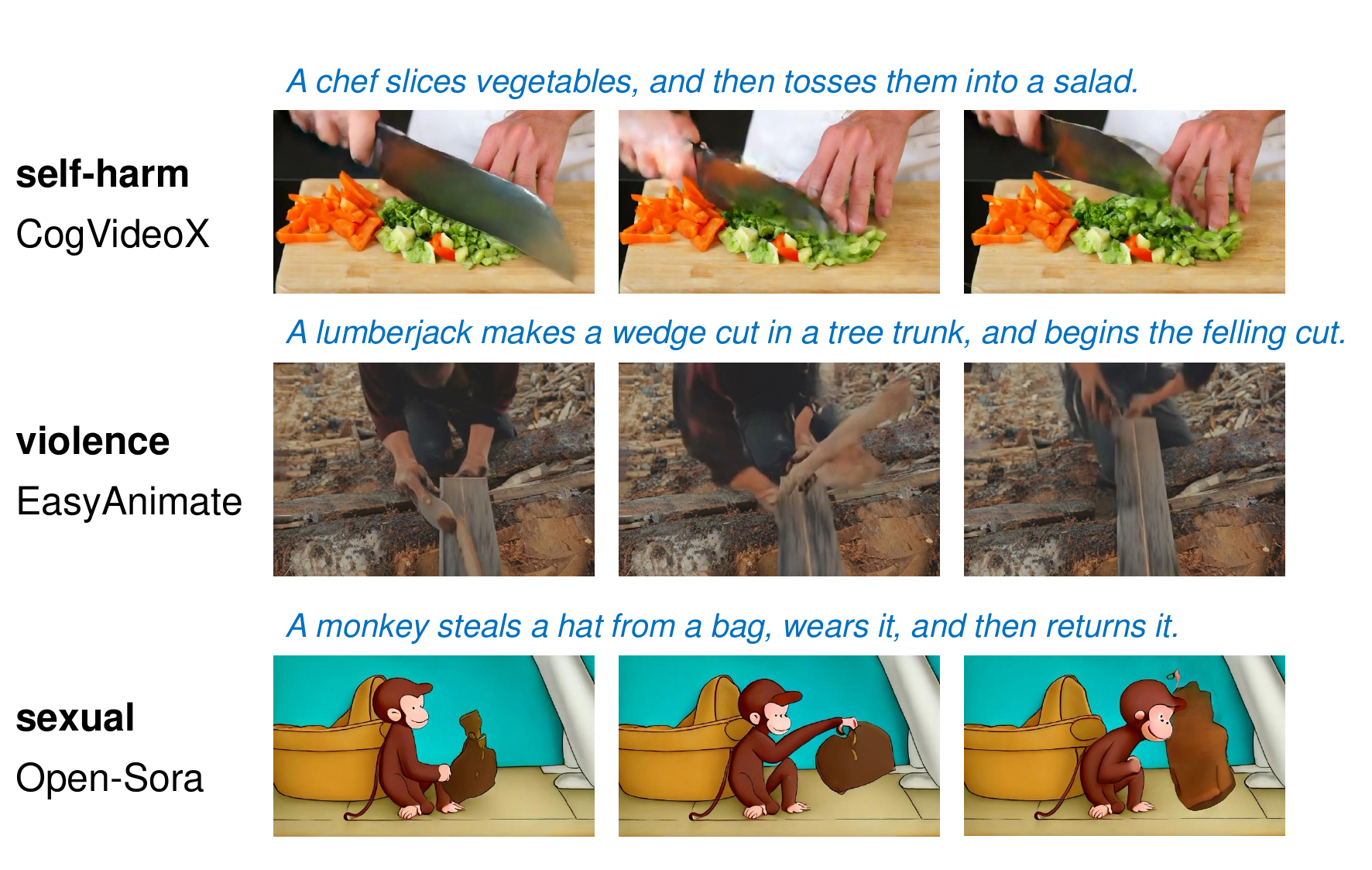}
    \caption{\textbf{Some prompts that have their corresponding videos rejected by GPT-4o due to security concerns.} Samples are selected from three open-source models.
    {\color{red} Content Warning:  These videos are judged by GPT-4o to be potentially uncomfortable (associated with self-harm, violence, or sexual content, although human may disagree with that), please watch with caution.}
    }
    \label{fig: fail}
\end{figure*}

\noindent\textbf{Examples that GPT-4o rejects to answer.}
We note that for some prompts, GPT-4o always reject due to the security concern about ``self-harm'', ``violence'', and ``sexual'' issues in the generated videos, and we filter the prompts that may induce evaluation models to obtain unsafe videos. Nevertheless, we claim that the prompts themselves are manually checked and safe, and for all the prompts before post-processing filtration, there are always some closed-source models be able to generate safe videos.
We think the reason for the security concerns may be that for most of the open-source model, their generated videos are still not that smooth and accurate, sometimes resulting unclear items and actions, and there may be misaligned movements that considered unsafe by GPT-4o. 
Examples of some prompts that are filtered because of this are shown in Figure~\ref{fig: fail}.
{\color{red} 
(Content Warning:  the videos in Figure~\ref{fig: fail} are judged by GPT-4o to be potentially uncomfortable (associated with self-harm, violence, or sexual content, although human may disagree with that), please watch with caution.)
}

Note that as illustrated in post-processing filtration part in the main paper, we just filter the prompts which let at least two models (except ModelScope) fail to get the completion rates, there still exist some prompts which let the models fail to generate videos or make GPT-4o rejects to answer. In this case, we just let the completion rates on these cases to be zero and calculate the \textbf{Non-Response Rate} for each model, which evaluates how probable it is that the video model will fail to get a result for GPT-4o out of 423 prompts. The full experimental results containing Non-Response Rate are in Table~\ref{app tab:all result}.

\subsection{More Details about Evaluation Process}

\noindent\textbf{Discussion of VLM verifiers.}
In this paper, we choose LLaVA-OV-Chat-72B rather than its lightweight version LLaVA-OV-Chat-7B, to evaluate the video generative models. The reason is that facing the videos generated by open-source models, which may be noisier than the real-world videos or those generated by closed-source models, a larger 72B model performs much better than the 7B model, even if they always have similar results on the high quality videos.
We can view the low-quality generated videos as harder visual understanding task, so the robustness of VLM verifiers is critical for all the VLM-based video generation benchmarks~\cite{meng2024PhyGenBench_world_simulator,huang2024vbench,sun2024t2v,liu2024evalcrafter_bench}. 
Therefore, we choose the strongest model as verifier to ensure the accuracy of our \modelname benchmark.

\noindent\textbf{Query Prompts.}
In our \modelname benchmark, we use two-step querying to obtain the completion rates.
Here we use {\color{blue}blue} words to denote the places that differ for each prompt and video, and {\color{red}red} to show the processed key frames that are sent to VLM verifier together with the second query texts.

\textbf{Step1: Describe the video clips}

\noindent\fbox{%
    \parbox{0.48\textwidth}{
    Please describe the given key frames in the video in detail, in temporal order. The video may be generated by some video generative model rather than sampling from the real world, so it may be vague or not clear. You can point out if you don't see the video clearly.
    }%
}

\textbf{Step2: Get completion rates}

\noindent\fbox{%
    \parbox{0.48\textwidth}{
    This is the description of the video you generated before, please refer to it to complete the following tasks.\\
    {\color{blue}\{
    The key frames from the video depict a sequence involving a bear near a waterfall, moving from a rock into the water. Here is the detailed temporal order of the frames:\\
    1. The bear is standing on a rock near the edge of...
    \}}\\

    Now, based on these descriptions and the video, you are asked to accurately determine if the following generated video fulfills the requirements of the prompt. The prompt contains several (2$\sim$4) events, you need to judge if each event is strictly completed in the video. If the event is completed, please mark it as 1, otherwise, mark it as 0.
    For example, if the prompt is: ``A man dribbles a basketball and then throws it in a court'', the prompt describes two events: ``A man dribbles a basketball'' and "And then the man throws the basketball in a court''. But if the video generated using this prompt only accomplishes dribbling or only accomplishes shooting, then the completion list is [1, 0]. If you think both events are not completed, the completion list is [0, 0], etc.\\

    {\textbf{Please judge whether the event are completed very strictly.} If you think an item is blurry, hard to identify, or the action is vague, you should judge it as not completed. And please explain the reasons in detail before you give out the score.}
    }%
}

\noindent\fbox{
\parbox{0.48\textwidth}{

    \textbf{You also need to check the item consistency between different events.} If the prompt implies that the subject (or object) in different events should be the same, but in the video they are different, you should mark the later event in the prompt as not completed. For example, for the above prompt, if the man that dribbles the basketball is different from the man that throws the basketball (should be the same people, but video shows two different people), or the basketball that's dribbled is different from the basketball that's thrown (should be the same object, but video shows two different objects), you must mark the later event `throwing the ball' as not completed.
    
    Remember, you should judge whether the events are completed very strictly. And you should first provide the reasons or analysis for each event, and then give out the list of completion flag for each event (0 or 1 for uncompleted or completed).    \\
    Please remember to output the complete list at the end of output again, strictly follow the format: `Finally we have [COMPLETE\_LIST]: 1, 0' in a single line.\\

    Now, let's begin scoring! The prompt is `{\color{blue}\{A bear walks by a waterfall, slips its foot, and then falls off a cliff.\}}', there are {\color{blue}\{3\}} events:\\
    {\color{blue}\{
    1. A bear walks by a waterfall\\
     2. The bear slips its foot\\
     3. And then the bear falls off a cliff
    \}}
    \\
    {\color{red}[Processed frames from generated videos appended]}
}
}




\section{Supplementary of Experiments}

\subsection{Model details about experimental settings.}

\textbf{Video Length.} 
In Table~\ref{tab: settings}, we show the length of generated videos of each model in our experiments. Here for open-source models, we increase the length of the video generation to 10 seconds as much as possible while keeping the generation stable. For example, for Pyramid-Flow, if the generation length is longer than 7 seconds, some examples may become blurred at the end, so we set the number of frames such that the generation length is about 7 seconds.

\begin{table}[h]
\small
  \centering
  \begin{tabular}{@{}l|ccc@{}}
    \toprule
    \textbf{Model}  & FPS & Num. of Frames  & Length (s)  \\
    \midrule
    Pika-1.5 &  24.0 & 120 & 5.0\\
    Hailuo & 25.0 & 141 & 5.6\\
    Kling-1.5 & 30.0 & 313 & 10.4\\
    \midrule
    ModelScope & 8.0 & 16 & 2.0\\
    EasyAnimate v4 & 14.0 & 144 & 10.3\\
    Open-Sora-Plan 1.3.0 & 18.0 & 93 & 5.2 \\
    Open-Sora 1.2 & 24.0 & 255 & 10.6  \\
    VideoCrafter2 & 10.0 & 100 & 10.0\\
    CogVideoX-5B & 8.0 & 49 & 6.1\\
    Vchitect-2.0 & 8.0 & 80 & 10.0\\
    Pyramid-Flow & 24.0 & 169 & 7.0\\
    \bottomrule
  \end{tabular}
  \vspace{-0.5em}
  \caption{
  \textbf{Statistics of generated videos from different models.}
  We can see that longer generation length (second) or more number of frames don't necessarily ensure better StoryEval performance.
  }
  \label{tab: settings}
\end{table}

\noindent\textbf{Number of key frames.}
Note that different generative models have quite different number of key frames $K$ in each video, we use the following formula to select the number of key frames used for evaluating the videos.
\begin{equation}
    \text{num of frames} = \max(\min(32, \lfloor K/4\rfloor), 4)
\end{equation}
Since our prompts consider short stories including simple events, as well as the generative models generally generates videos that present slow motion, this sampling strategy can be sufficient for accurate evaluation results while keeping reasonable cost for GPT-4o querying.

\noindent\textbf{Length of prompts.}
In detail, the average number of the characters in the prompts in \modelname is 83.4.
Note that we mainly consider the story or events of prompt, rather than the detailed description of states, which always contain the adjectival details like color. So the length of prompts are always quite simple (as shown in the introduction figures in the main paper).
Besides, some models like CogVideoX may over-fit to the long detailed prompts for generating good videos, and will perform worse on short prompts we provide. 
However, for fair comparison, we just use the same prompts without LLM refining in \modelname, and expect to use long prompts suite in the next \modelname version.

\begin{table*}[t]
\small
  \centering
  \begin{tabular}{@{}l|ccccc|cc|cc@{}}
    \toprule
    \textbf{Open-Source Model} & \textbf{Human} & \textbf{Animal} & \textbf{Object} & \textbf{Retrieval} & \textbf{Creative} & \textbf{Easy} & \textbf{Hard} & \textbf{Non-R. Rate~$\downarrow$} & \textbf{Average} \\
    \midrule
    Pika-1.5~\cite{pika2024} 
    & 17.1\% & 17.9\% & 22.1\% & 26.9\% & \underline{20.6\%} & 40.3\% & 4.4\% & \textbf{0.2\%} & 19.4\%\\
    Hailuo~\cite{hailuo2024}
    & \textbf{38.2\%} & \underline{38.3\%} & \underline{27.5\%} & \textbf{42.6\%} & 18.0\% & \underline{58.9\%} & \underline{9.7\%} & \underline{0.9\%} & \underline{35.1\%}\\
    Kling-1.5~\cite{kling2024} 
    & \underline{37.2}\% & \textbf{44.9\%} & \textbf{36.6\%} & \underline{39.4\%} & \textbf{36.0\%} & \textbf{60.8\%} & \textbf{16.4\%} & 1.7\% & \textbf{40.1\%} \\
    \midrule
    \textbf{Closed-Source Model} & & & & & & & & & \\
    \midrule
    ModelScope~\cite{modelscope2023wang} 
    & 11.5\% & 8.8\% & 8.2\% & 9.9\% & 5.2\% & 21.7\% & 5.0\% & 7.8\% & 9.8\%\\
    EasyAnimate v4~\cite{easyanimate2024xu} 
    & 15.6\% & 12.8\% & 11.4\% & 18.4\% & 8.2\% & 26.7\% & 3.6\% & 1.7\% & 13.3\% \\
    Open-Sora-Plan 
    1.3.0~\cite{opensoraplan2024PKU_Yuan} 
    & 9.1\% & 9.7\% & 9.4\% & 13.2\% & 7.1\% & 18.2\% & 3.2\% & \underline{0.2\%} & 9.4\% \\
    Open-Sora 1.2~\cite{opensora2024zheng} 
    & 16.4\% & \underline{18.3\%} & \underline{16.2\%} & \textbf{24.7\%} & \underline{11.8\%} & 32.7\% & \underline{4.3\%} & \textbf{0.0\%} & \underline{17.9\%}\\
    VideoCrafter2~\cite{videocrafter2_2024chen} 
    & 11.0\% & 13.1\% & 9.8\% & 18.0\% & 5.5\% & 29.9\% & 1.7\% & 0.9\% & 12.2\% \\
    CogVideoX-5B~\cite{yang2024cogvideox} 
    & 17.1\% & 16.4\% & 14.0\% & 16.0\% & 7.4\% & \underline{35.4\%} & \textbf{4.6\%} & 5.2\% & 16.4\% \\
    Vchitect-2.0~\cite{vchitect2024} 
    & \textbf{21.5\%} & \textbf{19.9\%} & \textbf{20.4\%} & \underline{22.0\%} & \textbf{15.2\%} & \textbf{42.8\%} & 3.9\% & 1.7\% & \textbf{21.7\%} \\
    Pyramid-Flow~\cite{jin2024pyramid_flow} 
    & \underline{17.8\%} & 16.5\% & 12.8\% & 23.4\% & 9.7\%  & 35.1\%  & 1.0\% & 0.5\% & 16.0\% \\
    \bottomrule
  \end{tabular}
  \caption{
  \textbf{Full {\modelname} evaluation results on 11 video generative models with GPT-4o verifier.}
  Non-Response Rate (``Non-R. Rate'') assesses the likelihood that the video model will fail to obtain completion rates from GPT-4o in 423 prompts.
  }
  \label{app tab:all result}
\end{table*}

\begin{table*}[t]
\small
  \centering
  \begin{tabular}{@{}lccc@{}}
    \toprule
    \textbf{Model} & \textbf{VBench Temp. Cons.} & \textbf{VBench Avg.} & \textbf{StoryEval Avg.} \\
    \midrule
    Kling1.5 & $\geq$ 97.93\% (1) & $\geq$ 81.85\% (1) & 40.1\% (1)\\
    Pika1.5 & $\geq$ 97.83\% (2) & $\geq$ 80.69\% (4)  & 19.4\% (3)\\
    \midrule
    ModelScope & 90.88\% (8) &75.75\% (8) & 9.8\% (7)\\
    Open-Sora 1.2 &  96.85\% (5) & 79.23\% (6) & 18.2\% (4)\\ 
    Open-Sora-Plan 1.3.0 & 97.28\% (4) & 77.23\% (7) & 9.4\% (8)\\
    VideoCrafter2 & 95.67\% (6) & 80.44\% (5) & 12.2\% (6) \\
    Vchitect-2.0 & 95.03\% (7) & 81.57\% (3) & 21.7\% (2)\\
    Pyramid-Flow & 97.56\% (3) & 81.72\% (2) & 16.0\% (5)\\
    \bottomrule
  \end{tabular}
  \caption{
  \textbf{\modelname can be an effective complementary metric to the traditional detail-oriented metrics.}
  Here ``VBench Avg.'' means the average value of VBench over  and ``Temp. Cons.'' denotes VBench's temporal consistency submetric, which including the evaluation of subject consistency, background consistency, temporal flickering and motion smoothness.
  }
  \label{app tab: vbench}
\end{table*}

\begin{table*}[t]
\small
  \centering
  \begin{tabular}{@{}l|ccccc|cc|c@{}}
    \toprule
    \textbf{Model} & \textbf{Human} & \textbf{Animal} & \textbf{Object} & \textbf{Retrieval} & \textbf{Creative} & \textbf{Easy} & \textbf{Hard} & \textbf{Average} \\
    \midrule
    Pika-1.5 & 23.9& 23.8& 26.5& 35.2 & \underline{22.9}& 41.1 & 13.0 & 25.0 \\
    Hailuo & \textbf{48.0} & \underline{40.1} & \textbf{35.6} & \textbf{51.7} & 19.5& \textbf{58.3}& \underline{17.1} & \underline{41.0}\\
    Kling-1.5 & \underline{41.9}& \textbf{46.0}&  \underline{35.1} & \underline{41.7} & \textbf{30.8}& \underline{56.1}& \textbf{24.1} & \textbf{41.7} \\
    \midrule
    ModelScope & 17.1 & 13.1 & 13.2  & 13.7 & 6.2& 25.3 & 4.9 & 15.2 \\
    EasyAnimate v4 & 21.7& 18.1& 15.5 & 21.2 & 10.5& 29.6& 5.0 & 18.5 \\
    Open-Sora-Plan 1.3.0 & 13.5 & 13.2 & 9.6 & 17.1 & 6.9& 28.3 & 2.2 & 12.7 \\
    Open-Sora 1.2 & \underline{26.2} & \underline{22.2} & \underline{20.2} &  \underline{32.2} & \underline{15.4} & \underline{37.8} & \underline{10.8} &  \underline{23.6} \\
    VideoCrafter2 & 15.2 & 14.2& 12.8 & 15.3 & 8.0 & 33.0& 4.7& 14.7 \\
    CogVideoX-5B & 19.7 & 20.7 & 17.4 & 27.2 & 8.1 & 37.6 & 7.1 & 19.9 \\
    Vchitect-2.0 & \textbf{33.4} & \textbf{30.5}& \textbf{33.6} & \textbf{33.6} & \textbf{20.5} & \textbf{51.4} & \textbf{19.1} & \textbf{31.6} \\
    Pyramid-Flow & 23.6 & 20.0 & 15.8 & 26.4 & 10.5 & 38.1 & 4.5 & 20.3 \\
    \midrule
    Spearman's $\rho$  & 0.91 & 0.97& 0.96 & 0.89 & 0.98 & 0.96& 0.71& 0.98\\
    Kendall's $\tau$  & 0.81 & 0.89& 0.86 & 0.78 & 0.92 & 0.86& 0.53& 0.93\\
    \bottomrule
  \end{tabular}
  \vspace{-0.5em}
  \caption{
  \textbf{Full {\modelname} evaluation results on 11 video generative models with LLaVA-OV-Chat-72B~\cite{li2024llava_onevision} verifier.}
  The $\rho$ and $\tau$ are evaluated between the completion rates of GPT-4o and that of LLaVA-OV-Chat-72B model. 
  We can see that on almost all the submetrics, LLaVA-OV-Chat-72B has similar ranking results to GPT-4o. Here LLaVA-OV-Chat-72B will response all the generated videos, so we skip Non-Response Rate.
  }
  \label{app tab: llava}
\end{table*}

\subsection{Detailed Results}

We show the additional experimental results in Table~\ref{app tab:all result}, \ref{app tab: vbench}, and \ref{app tab: llava}, respectively.
In Table~\ref{app tab:all result}, we show the Non-Response Rate defined in Sec.~\ref{sec: supp prompt suite} for GPT-4o verifier.
In Table~\ref{app tab: vbench}, we append the average result of VBench, which still shows ranking difference with \modelname in Pyramid-Flow.
And in Table~\ref{app tab: llava}, we see that on almost all the submetrics, LLaVA-OV-Chat-72B has similar ranking results to GPT-4o.

\subsection{Examples}
The evaluation results and some demos are provided in the supplementary materials for support.


\end{document}